\pdfoutput=1

\documentclass[11pt]{article}

\usepackage[final]{acl}

\usepackage{times}
\usepackage{latexsym}

\usepackage[T1]{fontenc}

\usepackage[utf8]{inputenc}

\usepackage{microtype}

\usepackage{inconsolata}

\usepackage{bm}
\usepackage{tikzducks}
\usepackage{graphicx}
\usepackage[most]{tcolorbox}
\usepackage{colortbl}
\usepackage{pifont}
\usepackage{enumitem}
\definecolor{Gray}{gray}{0.9}

\usepackage{listings}
\definecolor{codeblue}{rgb}{0.25,0.5,0.5}
\definecolor{codekw}{rgb}{0.85, 0.18, 0.50}
\lstdefinestyle{mystyle}{
    backgroundcolor=\color{white},
    basicstyle=\fontsize{7.5pt}{7.5pt}\ttfamily\selectfont,
    columns=fullflexible,
    breaklines=true,
    captionpos=b,
    commentstyle=\fontsize{7.5pt}{7.5pt}\color{codeblue},
    keywordstyle=\fontsize{7.5pt}{7.5pt}\color{codekw},
}
\usepackage{graphicx}
\usepackage{amsmath, amssymb, bm}
\usepackage{algorithm}
\usepackage{multirow}
\usepackage{graphicx}
\usepackage{booktabs}
\usepackage{multicol}
\usepackage{hyperref}
\usepackage{amsmath}
\usepackage{caption}
\usepackage{enumitem}
\usepackage{wrapfig}
\usepackage{listings}
\usepackage{pifont}
\usepackage{array}
\usepackage{placeins}
\usepackage{float}
\definecolor{codegreen}{rgb}{0.0, 0.411, 0.243}
\definecolor{codered}{rgb}{0.89, 0.26, 0.20}
\usepackage{hyperref}
\definecolor{dartgreen}{HTML}{00693e}
\definecolor{refcolor}{HTML}{9F363A}
\hypersetup{
    colorlinks=true,
    linkcolor=refcolor,
    citecolor=refcolor,
    filecolor=magenta,      
    urlcolor=refcolor,
    }

\title{Temporal Working Memory: Query-Guided Segment Refinement for Enhanced Multimodal Understanding}
\author{
 \textbf{Xingjian Diao$^{*}$},
 \textbf{Chunhui Zhang$^{*}$},
 \textbf{Weiyi Wu},
 \textbf{Zhongyu Ouyang},
\\
 \textbf{Peijun Qing},
 \textbf{Ming Cheng},
 \textbf{Soroush Vosoughi},
 \textbf{Jiang Gui}
\\
 Dartmouth College
 \\
   \texttt{\{xingjian.diao, chunhui.zhang, weiyi.wu\}.gr@dartmouth.edu}
 \\
   \texttt{\{soroush.vosoughi, jiang.gui\}@dartmouth.edu}
}
\begin{document}
\maketitle

\newcommand\blfootnote[1]{%
  \begingroup
  \renewcommand\thefootnote{}\footnote{#1}%
  \addtocounter{footnote}{-1}%
  \endgroup
}

\blfootnote{$^{*}$contributed equally}

\begin{abstract}
Multimodal foundation models (MFMs) have demonstrated significant success in tasks such as visual captioning, question answering, and image-text retrieval. However, these models face inherent limitations due to their finite internal capacity, which restricts their ability to process extended temporal sequences---an essential requirement for comprehensive video and audio analysis. 
To overcome these challenges, we introduce a specialized cognitive module, temporal working memory (TWM), which aims to enhance the temporal modeling capabilities of MFMs. It selectively retains task-relevant information across temporal dimensions, ensuring that critical details are preserved throughout the processing of video and audio content. The TWM uses a query-guided attention approach to focus on the most informative multimodal segments within temporal sequences. By retaining only the most relevant content, TWM optimizes the use of the model's limited capacity, enhancing its temporal modeling ability.
This plug-and-play module can be easily integrated into existing MFMs. With our TWM, nine state-of-the-art models exhibit significant performance improvements across tasks such as video captioning, question answering, and video-text retrieval. By enhancing temporal modeling, TWM extends the capability of MFMs to handle complex, time-sensitive data effectively. Our code is available at \url{https://github.com/xid32/NAACL_2025_TWM}.
\end{abstract}

\section{Introduction}
\label{introduction}

Multimodal foundation models (MFMs) have demonstrated impressive capabilities in tasks such as video captioning, question-answering, image-text retrieval, and broader multimodal understanding \cite{zhang2022look, kim2024show, he2024ma, jian2024expedited, zhang2025pretrained, yao2024customized, xie2024uncertainty, yao2024multi, xie2025multi, Han_2024_CVPR, liu2024protecting,  lin2024learning}. While MFMs excel at processing multimodal inputs, MFMs are often not equipped to explicitly reduce the input context burden, particularly in extracting query-relevant information from the input context for video understanding tasks. 

\begin{figure}[tb]
\centering
\resizebox{0.48\textwidth}{!}{
\includegraphics{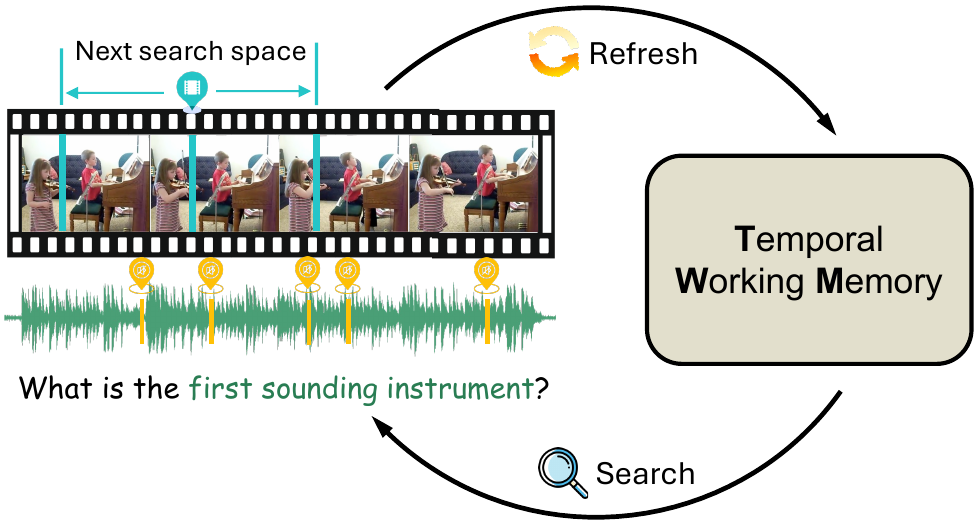}
}
\caption{Temporal Working Memory (TWM): TWM employs search engine and memory refresh mechanisms to retain key segments in long multimodal inputs.}
\label{fig:teaser}
\vspace{-0.6cm}
\end{figure}

In humans, working memory retains and processes information over short time spans with limited capacity \cite{baddeley2000episodic}, and similar constraints apply to MFMs \cite{liu2024lost}. For example, LLaMA has a context length limit of 2048 tokens \cite{touvron2023llama1, touvron2023llama2, dubey2024llama3}, while LLaVA \cite{Liu2023Vis} and BLIP-2 \cite{pmlr-v202-li23q} can handle only 256 and 32 tokens per image, respectively. 
These limited capacities prevent models from effectively retaining sufficient information over extended temporal spans.

While operating under similar constraints, the human cognitive system has evolved effective mechanisms for efficient information processing \cite{baddeley2000episodic}, such as selective retention and distraction filtering, to dynamically prioritize relevant information while discarding irrelevant details ~\cite{zhangchunhui2024working, gong2024working}.
In contrast, current MFMs lack such selective mechanisms found in human working memory, preventing them from effectively filtering and retaining the most relevant temporal segments from multimodal inputs (e.g., video frames or audio clips). Consequently,  MFMs tend to process the entire input sequences indiscriminately within their perceptual window, leading to inefficient utilization of the model’s capabilities.

Recent developments in the memory of LLMs have significantly improved their ability to manage temporal contexts. Various approaches have been proposed to address the inherent memory limitations of LLMs and thereby improve their performance on complex tasks.
\citet{li-etal-2023-large} proposed a knowledge-aware fine-tuning method that instructs LLMs to prioritize relevant external context while minimizing reliance on internal pre-trained knowledge, thereby enhancing their memory by filtering out irrelevant information. Building on this, \citet{gong2024working} demonstrated that ChatGPT's working memory is similar to that of humans, suggesting that enhancing memory capacity is crucial for advancing the intelligence of AI systems. Further exploring these limitations, \citet{zhangchunhui2024working} recommended strategies for more efficient memory utilization, highlighting the importance of improving both memory retention and model autonomy for better reasoning capabilities. In the context of multimodal models, \citet{wu2024v} integrated visual working memory mechanisms to help models focus on essential features in high-resolution images, significantly improving visual grounding performance.

Despite recent advances in LLM memory management, these models still face fundamental challenges in dynamic multimodal and temporal reasoning: 
\textit{(i)}
they lack effective mechanisms to filter and retain query-relevant information from multimodal inputs (e.g., audio, video, language), leading to indiscriminate processing of entire sequences regardless of their relevance.
\textit{(ii)} these models lack the ability to effectively capture temporal dependencies. They struggle with short-term changes (e.g., rapid changes in audio or visual content) and long-term relationships (e.g., understanding how earlier events relate to later ones). This reduces their ability to reason about sequential and time-sensitive data, which is crucial for tasks involving events that unfold over time.
\textit{(iii)} the models' inability to efficiently process large volumes of raw data leads to information overload, straining their limited capacity and degrading performance when dealing with complex and high-volume data.

Therefore, we draw on human working memory to effectively \textbf{extract query-relevant multimodal information across temporal dimensions}. We propose a multimodal Temporal Working Memory (TWM) mechanism for MFMs, as shown in Figure \ref{fig:teaser}. This mechanism employs a query-guided attention mechanism to selectively retain only query-relevant audio and visual inputs, focusing on the most informative segments along the temporal axis. The TWM mechanism constructs a temporal memory buffer at the model input stage, enabling MFMs to efficiently store and manage critical information across time. By concentrating on the retention of the most relevant data, TWM significantly improves the model's ability to reason over extended temporal sequences in a multimodal context. Our contributions are:
\begin{itemize}[leftmargin=*]
\item We propose a Temporal Working Memory (TWM) mechanism with an integrated query-guided selection module. This module directs the model to retain key segments in long video and audio sequences, optimizing the use of the model's limited capacity.

\item We employ a multi-scale temporal attention mechanism for both local and global dependencies, enabling accurate identification of relevant video-audio segments across temporal inputs.

\item We integrate TWM into nine state-of-the-art MFMs and evaluate our approach on three large-scale multimodal benchmarks, covering tasks including audio-visual question answering, video captioning, and video-text retrieval. Our approach effectively yields significant performance improvements across all tasks.
\end{itemize}

\section{Related Works}
\label{related_works}
\paragraph{Temporal Modeling in MLLMs}
Multimodal LLMs (MLLMs) for long-video understanding aim to capture long-range temporal patterns. A common strategy is temporal pooling, as used in VideoChatGPT \cite{maaz2023video}, but this can limit performance due to inadequate temporal modeling. More advanced methods, such as video-LLAMA \cite{zhang2023video}, incorporate video query transformers to enhance temporal dynamics, but this comes with increased model complexity. To reduce computational demands, some models rely on pre-extracted features, avoiding joint training of backbone architectures \cite{hussein2019timeception, liu2024breaking, wu2021towards}. Techniques like Vis4mer \cite{islam2022long} and S5 \cite{wang2023selective} utilize the S4 transformer architecture \cite{gu2021efficiently} for efficient long-range temporal modeling. Recent developments, such as online video processing \cite{he2024ma}, employ memory banks to track past content for long-term analysis. In contrast, we propose a TWM mechanism that retains only query-relevant multimodal inputs through search engines within a temporal context.

\paragraph{Video Understanding}
Video understanding tasks evaluate a model's ability to process multimodal content, focusing on both temporal and semantic aspects. Key tasks for long-video understanding include audio-visual question answering (AVQA), video captioning, and video-text retrieval, supported by extensive research and large-scale datasets \cite{liu2024tackling, xu2016msr, bain2020condensed}. Prior AVQA methods fine-tuned pretrained visual models with adapters \cite{liu2024tackling, Lin_2023_CVPR, NEURIPS2023_af01716e, diao2024learning}, while recent approaches use unified multimodal encoders with LLMs \cite{Han_2024_CVPR}. Video captioning models employ graph neural networks (GNNs) \cite{hendria2023action}, simplified image-text architectures \cite{wang2022git}, or causal effect networks (CEN) to enhance temporal coherence \cite{nadeem2024narrativebridge}. In video-text retrieval, adaptive frame aggregation reduces visual tokens to accelerate encoding \cite{ren2023testa}.
In contrast to previous work focusing on specific multimodal applications, this study emphasizes the role of TWM in enhancing fundamental temporal grounding across audio, video, and language.

\section{Temporal Working Memory}
\label{method}

\begin{figure*}[tb]
\begin{center}
\includegraphics[width=1\linewidth]{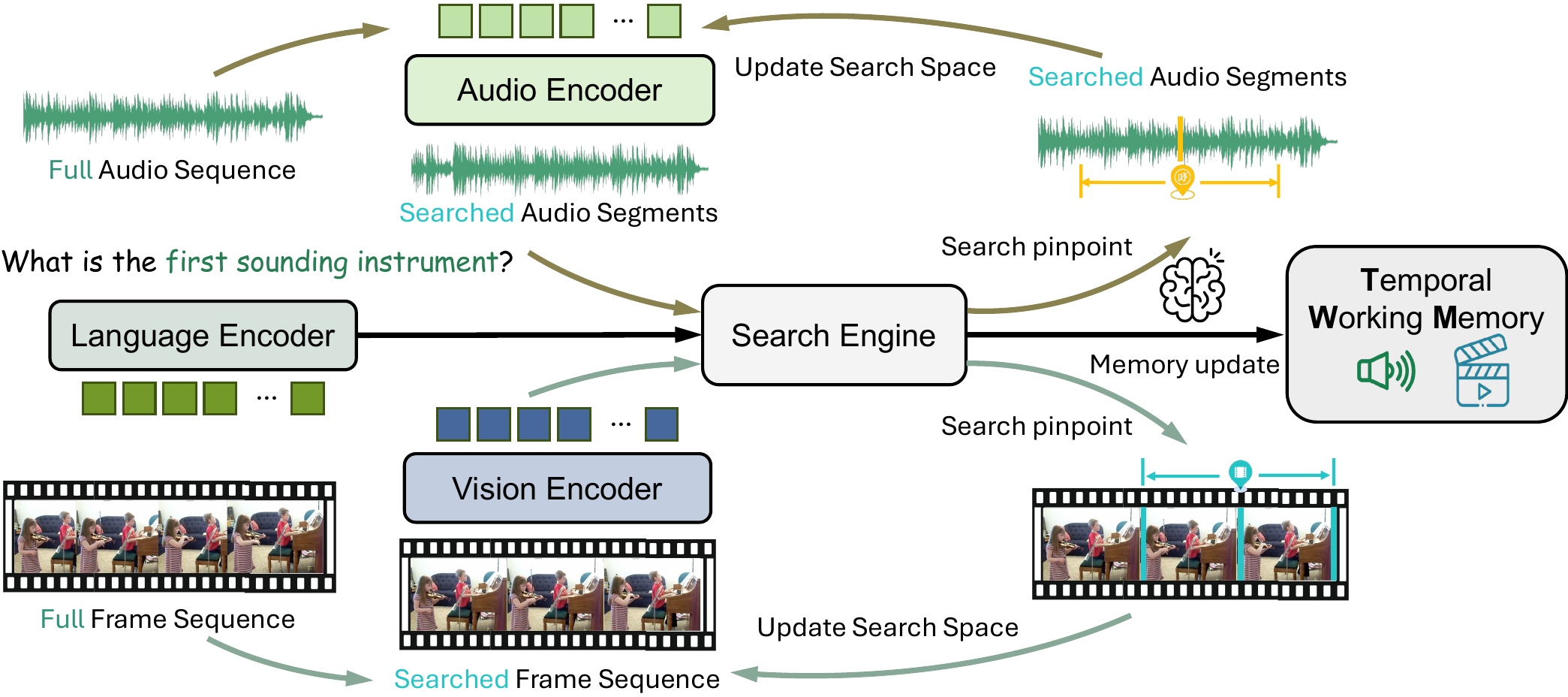}
\end{center}
\caption{The temporal working memory (TWM) pipeline retains the most relevant segments from video and audio inputs based on a language query. The Language Encoder processes the query, guiding the Search Engine to identify and select key video and audio segments. TWM ensures the retention of only the most informative data, enabling the efficient utilization of multimodal foundation models' capabilities.}
\label{fig:pipeline}
\end{figure*}

Our temporal working memory (TWM) framework is a complementary architecture for multimodal large language models, integrating visual and auditory buffer components. If the model does not involve audio, the auditory component can be omitted.  The working pipeline of the TWM framework is outlined in Algorithm \ref{algo:TWM}, while the search and memory update processes are depicted in Figure \ref{fig:pipeline}. 

\begin{algorithm}[t]
    \caption{\normalsize TWM for MFMs}
    \label{algo:TWM}
    \lstset{
      basicstyle=\fontsize{9pt}{9pt}\ttfamily, 
      columns=fullflexible,
      breaklines=true,
      captionpos=b
    }
    \begin{lstlisting}[language=python]
# search query-related segments from video&audio
def neural_search(video, audio, qry):
    # Step 1: Segment and encode video and audio
    v_seg, a_seg = segment(video, audio)
    v_embs, a_embs = encode(v_seg, a_seg)
    
    # Step 2: Calculate relevance scores
    v_scores, a_scores = sim(v_embs, a_embs, qry)
    
    # Step 3: Iterate to select the relevant segments
    v_buffer, a_buffer= select(v_embs, a_embs, v_scores, a_scores)
    
    return v_buffer, a_buffer
    
# temporal working memory for video and audio
v_buffer, a_buffer = neural_search(video, audio, qry)

# MFM with temporal working memory
output = MFM(v_buffer, a_buffer, qry)
    \end{lstlisting}
\end{algorithm}

\subsection{Visual Memory Management}
\subsubsection{Query-Relevant Video-Frame Search}
\label{Query-Relevant Video-Frame Selection}
Our mechanism emulates human cognitive strategies by identifying and retaining critical visual information from long video sequences. It alternates between two key operations—\textbf{search} and \textbf{update}—to dynamically adjust memory and focus on the most relevant segments, as shown in Figure~\ref{fig:pipeline}.

Initially, $k$ frames are selected from the full sequence of $N$ frames and processed through a visual encoder to generate embeddings. Each frame $v_i$ is assigned a \textbf{Similarity Score} ($S(v_i)$), defined as:
\begin{equation}
S(v_i) = \alpha_{1} D(v_i) + \alpha_{2} R(v_i, q),
\end{equation}
where $D(v_i)$ is function representing the \textbf{distinctiveness} of frame $v_i$, $R(v_i, q)$ is function representing the \textbf{relevance} of frame $v_i$ to the query $q$. $\alpha_{1}$ and $\alpha_{2}$ are adaptive weights that vary based on the nature of the downstream task and the duration of the video samples.
The selection process is iterative. In each iteration, the frame with the highest similarity score ($S(v_i)$) is chosen as the midpoint, and $k$ frames are searched uniformly within a range of $\frac{N}{k}$ around it. These frames are added to visual memory, excluding frames already present to maintain diversity. The process concludes upon convergence.

\subsubsection{Training Neural Search Engine}
\begin{figure}[tb]
\centering
\resizebox{0.3\textwidth}{!}{
\includegraphics{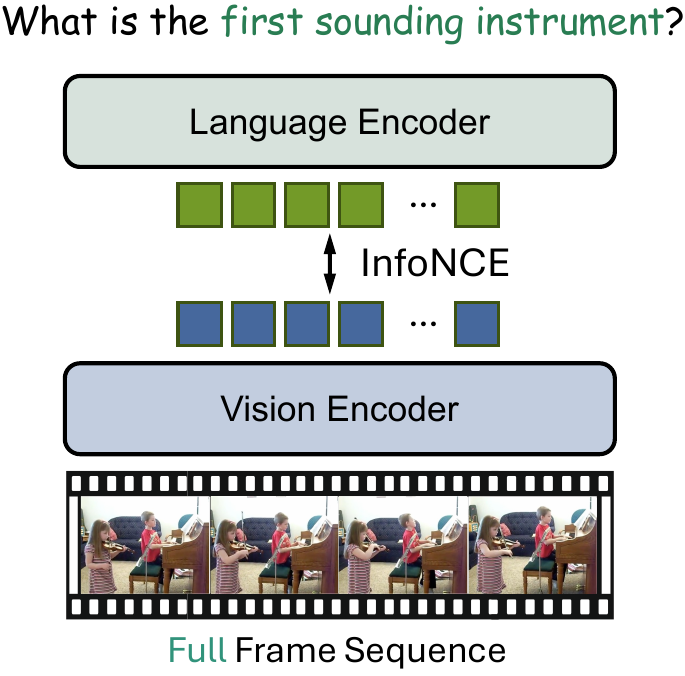}
}
\caption{Aligning frames with language query. A linear projection layer trained with InfoNCE loss aligns visual embeddings with query-based anchors.}
\label{fig:v-l_align}
\vspace{-0.4cm}
\end{figure}

To identify frames relevant to a given query, we use a \textbf{cross-modal alignment strategy} (Figure~\ref{fig:v-l_align}). Pretrained visual and language encoders are employed, with a linear projection layer that maps visual embeddings into the textual embedding space. The \textbf{InfoNCE loss}~\cite{oord2018representation} is used to optimize this alignment:
\begin{equation}
\mathcal{L}_{\text{InfoNCE}} = - \log \frac{\exp\left(\frac{\text{sim}(\mathbf{e}_{v}, \mathbf{e}_{t_i})}{\tau}\right)}{\sum_{j=1}^{N} \exp\left(\frac{\text{sim}(\mathbf{e}_v, \mathbf{e}_{t_j})}{\tau}\right)},
\end{equation}
where $\mathbf{e}_{v}$ is the embedding of video frame $v$, $\mathbf{e}_{t_i}$ is the embedding of text description $t_i$, $\text{sim}(x, y)$ denotes the cosine similarity function, $\tau$ is the temperature parameter controlling the sharpness of the distribution, and $N$ represents the number of negative samples.
The {InfoNCE loss} maximizes the similarity between the corresponding video and text embeddings while minimizing the similarity to unrelated samples. This ensures that the model effectively aligns the most relevant frames with their corresponding text, thereby optimizing its ability to retain meaningful visual information.

\subsection{Auditory Memory Management}
\subsubsection{Query-Relevant Audio-Segment Search}
The search process for identifying key audio segments mirrors the methodology used for video frames. The audio sequence is divided into predefined segments, typically 5-6 segments depending on video length, to model adequate temporal dependencies for tasks like AVQA and video captioning. This segmentation allows the model to focus efficiently on relevant audio intervals, improving attention allocation across extended sequences.

\begin{figure}[tb]
\centering
\resizebox{0.4\textwidth}{!}{
\includegraphics{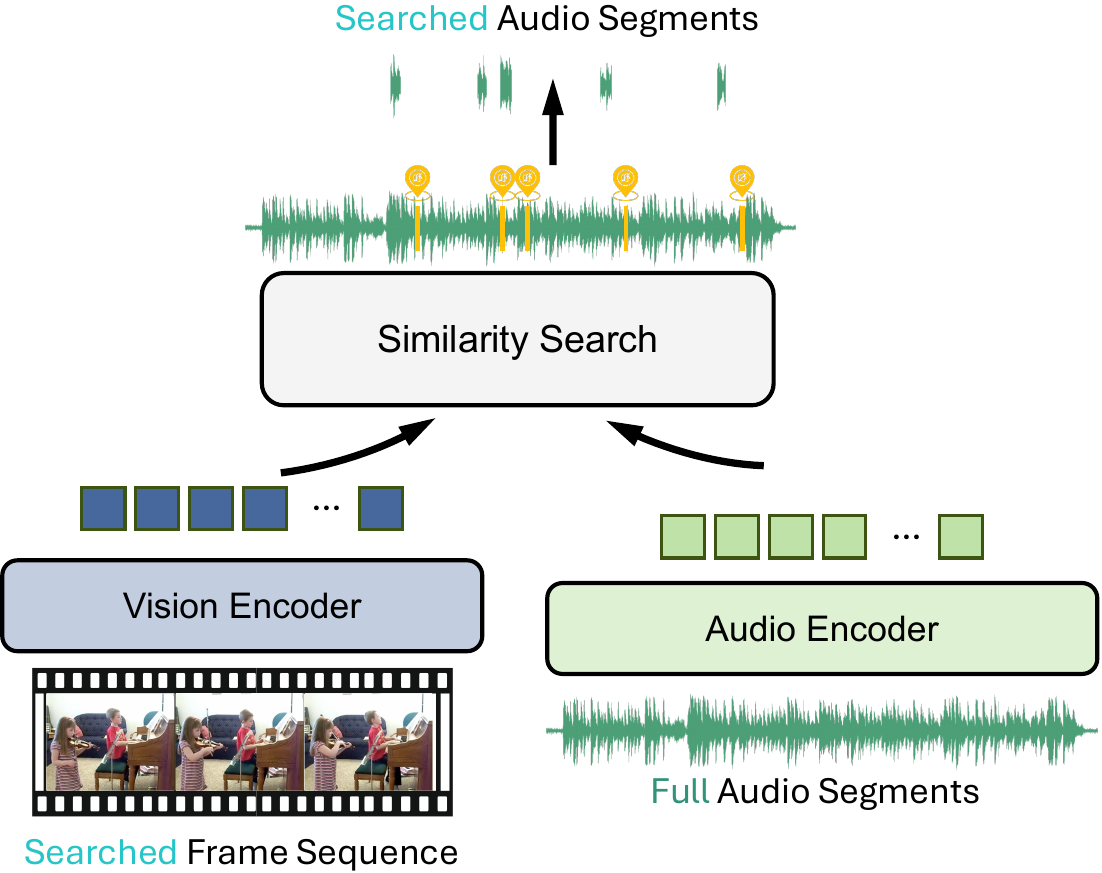}
}
\caption{Similarity search for query-relevant audio segments. The audio encoder utilizes visual embeddings as a query to search for the most relevant audio segments, updating the auditory buffer to retain only the essential audio information.}
\label{fig:audio_search}
\vspace{-0.5cm}
\end{figure}

Building upon Pathformer's dual-attention mechanism \cite{chen2024pathformer}, we extend their approach to enhance the correlation between audio and video data. Specifically, visual embeddings are used as queries in both attentions (Figure \ref{fig:audio_search}). To enable multimodal audio-visual synchronization, we integrate audio patches derived from Mel-spectrograms for temporal segmentation. This facilitates the alignment between the audio and visual inputs. Our audio encoder employs two key attention mechanisms:
\begin{itemize}[leftmargin=*]
    \item {Inter-segment attention} is designed to model \textit{global dependencies} across audio segments, enabling the model to capture broader relationships such as shifts in tone, mood, or overall sound context. Specifically, inter-segment attention calculates attention scores between the query $Q$, which is derived from visual features, and the keys $K_i$ from the audio segments $Att_{\text{inter}} = \text{softmax}\left(\frac{Q K_i^T}{\sqrt{d_K}}\right) V_i, \quad i \in [1, n].$ 
    Here, $Q$ represents the visual embeddings, while $K_i$ and $V_i$ are the audio embeddings from segment $i$. By using visual features as the query, this attention aligns audio information with relevant visual cues, effectively capturing how the audio context evolves to the video over time.

    \item {Intra-segment attention} aims to capture \textit{local dependencies} within individual audio segments, thereby modeling fine-grained temporal patterns such as audio variations or sudden changes in sound effects. The result of intra-segment attention for each segment is computed as: $Att_{\text{intra}} = \text{Concat}(Att_{\text{intra}_i} \mid i = 1, \dots, n).$  $Att_{\text{intra}_i}$ represents the computed attention within each segment $i$. The concatenation operation aggregates these intra-segment attention results across all segments, ensuring that short-term changes are captured and preserved for subsequent processing. This aggregation allows the model to retain a detailed representation of the short-term dynamics within each segment.
\end{itemize}

In the fusion layer, we apply a cross-modal attention mechanism to synchronize features from both audio and visual streams. Additionally, multi-kernel pooling aggregates audio patches across different-scale temporal dependencies, enhancing the alignment and understanding of temporal multimodal inputs.

As shown in Figure \ref{fig:v-a_align}, a pretrained visual feature extractor is used to align the audio segments with their corresponding visual frames to establish cross-modal coherence. To identify query-relevant audio patches, we apply cosine similarity between audio embeddings ($\mathbf{e}_{a_i}$) and visual embeddings ($\mathbf{e}_{v}$) as
$\text{sim}(\mathbf{e}_{a_i}, \mathbf{e}_{v}) = \frac{\mathbf{e}_{a_i} \cdot \mathbf{e}_{v}}{\|\mathbf{e}_{a_i}\| \|\mathbf{e}_{v}\|}$.
This similarity score is used to select the audio patches that are most relevant to the corresponding visual frames. The selected audio segments are then updated in the auditory buffer, ensuring that the most important audio information is retained. This iterative refinement enhances the synchronization and complementarity between audio and visual content.

\subsubsection{Training Audio Search Engine}
\begin{figure}[tb]
\centering
\resizebox{0.25\textwidth}{!}{
\includegraphics{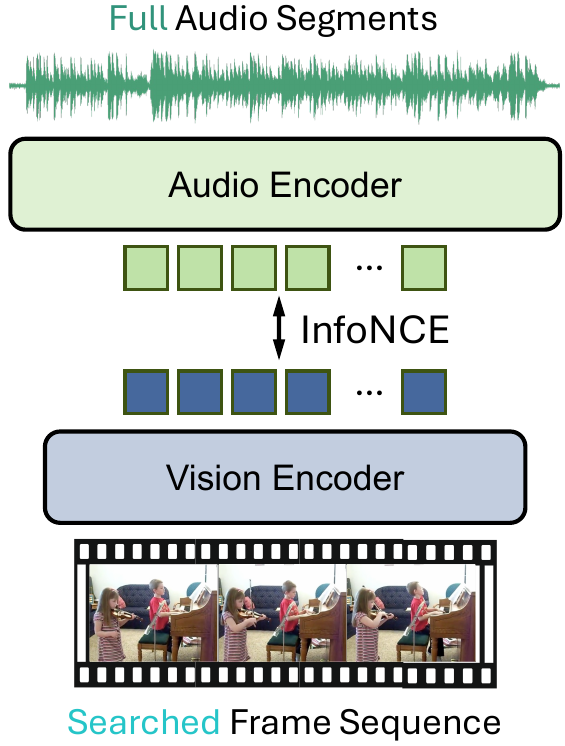}
}
\caption{Audio segments aligned with query-relevant frames. An audio encoder learns temporal distance and resolution for audio-visual embedding alignment.}
\label{fig:v-a_align}
\vspace{-0.3cm}
\end{figure}

To identify query-relevant audio segments, we also use InfoNCE loss to achieve cross-modal alignment (see Figure \ref{fig:v-a_align}). Let $\mathbf{e}_{a_i}$ denote the embedding of the audio patch $a_i$, and let $\mathbf{e}_{v}$ denote the embedding of the video frame $v$. The alignment loss is:
\begin{equation}
\mathcal{L}_{\text{InfoNCE}} = - \log \frac{\exp\left(\frac{\text{sim}(\mathbf{e}_v, \mathbf{e}_{a_i})}{\tau} \right)}{\sum_{j=1}^{N} \exp\left(\frac{\text{sim}(\mathbf{e}_v, \mathbf{e}_{a_j})}{\tau} \right)},
\end{equation}
where $\tau$ is a temperature parameter controlling the distribution's sharpness, and $N$ is the number of negative samples. This approach ensures effective alignment between audio and visual embeddings, allowing the model to identify cross-modal relationships effectively and refine the auditory buffer.

\section{Experiments}
\label{experiments}
We perform three major experiments to validate the effectiveness of the Temporal Working Memory mechanism discussed in the previous section. We evaluated the performance of state-of-the-art baseline models and the same models augmented with our Temporal Working Memory mechanism on the following downstream tasks: (1) \textbf{audio-visual question answering} (AVQA), (2) \textbf{video captioning}, and (3) \textbf{video-text retrieval}.

\subsection{Setup}
\label{setup}
\paragraph{Datasets}
We conduct experiments on AVQA, video captioning, and video-text retrieval:
\begin{itemize}[leftmargin=*,noitemsep,nolistsep] \item \textbf{MUSIC-AVQA v2.0} \cite{liu2024tackling}: MUSIC-AVQA v2.0 introduces 1,230 additional videos and 8,100 new question-answer (QA) pairs to further mitigate data bias. It builds on the original MUSIC-AVQA dataset \cite{Li2022learning}, which contains 9,288 videos of 22 musical instruments (over 150 hours) with 45,867 question-answer (QA) pairs across 33 templates in 9 categories. \item \textbf{MSR-VTT} \cite{xu2016msr}: Comprises 10,000 video clips (over 41 hours) from various online sources. Each video has 20 human-annotated captions, totaling 200,000 video-text pairs across diverse categories like music, sports, and gaming. \item \textbf{CMD} \cite{bain2020condensed}: The Condensed Movies Dataset includes over 33,000 clips from 3,600 movies, averaging 2 minutes each, with annotations such as captions, dialogues, and action labels, ideal for video-text retrieval tasks. \end{itemize}

\paragraph{Baselines}
Details of the baseline models can be found in Appendix \ref{baselines}. We evaluate nine state-of-the-art MFMs reproduced with open-source code and pretrained weights.

\paragraph{Evaluation Metrics}
Below standard metrics reflect the accuracy, quality, and retrieval capabilities:
\begin{itemize}[leftmargin=*,noitemsep,nolistsep] \item \textbf{Audio-Visual Question Answering}: Accuracy is measured for Audio (Counting, Comparative), Visual (Counting, Location), Audio-Visual (Existential, Counting, Location, Comparative, Temporal) question types, along with average accuracies for Audio, Visual, Audio-Visual, and overall. \item \textbf{Video Captioning}: Metrics include ROUGE-L, CIDEr, and SPICE, assessing overlap with ground truth, consensus with human annotations, and diversity of scene descriptions, respectively. \item \textbf{Video-Text Retrieval}: Metrics are Recall@1, Recall@5, and Recall@10, measuring retrieval performance within top 1, 5, and 10 predictions. \end{itemize}

\paragraph{Implementations}
The settings of each dataset:
\begin{itemize}[leftmargin=*,noitemsep,nolistsep] \item \textbf{MUSIC-AVQA v2.0}: Videos are 60 seconds long, with questions targeting specific portions of the video. We set $k = 11$ and ran 6 iterations, using $\alpha_{1} = 0.2$ and $\alpha_{2} = 0.8$, resulting in frame sampling rates consistent with 1 frame per second (fps). Audio segments are extracted every 5 seconds, selecting the highest-scoring segment from a total of 12 segments.
\item \textbf{MSR-VTT}: With a frame rate of 21 fps, we set $k = 3$ and ran 3 iterations, yielding 8--9 searched frames, with $\alpha_{1} = 0.5$ and $\alpha_{2} = 0.5$ for balanced frame selection.

\item \textbf{CMD}: With a frame rate of 30 fps, we used $k = 5$ and ran 7 iterations, producing 30--35 frames in total, using $\alpha_{1} = 0.6$ and $\alpha_{2} = 0.4$ to prioritize frame diversity.
\end{itemize}

As depicted in Figure ~\ref{fig:v-l_align}, the dimensions of the linear mapping layer, and similarly in Figure ~\ref{fig:v-a_align} for the fusion layer, correspond to the output embedding dimensions of the text encoder used by each model. The embedding dimensions typically span 768-D, 4096-D, or 16384-D, as detailed in Section~\ref{setup} where the baseline models are referenced.

Training is conducted on the NVIDIA H100 80GB GPUs using PyTorch. The Adam optimizer with a learning rate of $1e^{-4}$ is used, and each model is trained for 10 epochs.  On the MUSIC-AVQA dataset, each model requires an average training time of 65.2 hours. For the MSR-VTT dataset, the average training time per model is 14.6 hours, while for the CMD dataset, each model takes approximately 146.6 GPU-hours to train.

\subsection{Overall Comparisons}
\subsubsection{Audio-Visual Question Answering}
\paragraph{TWM captures fine-grained multimodal dependencies for comparative reasoning} 
In AVQA (Table \ref{tab:tab1}), TWM excels in identifying and preserving fine-grained dependencies between audio and visual inputs, especially in complex comparative tasks. In audio-related comparative QA, LAVisH+TWM improves by 12.40\%, DG-SCT+TWM gains 10.12\% and LAST-Att+TWM shows an increase of 11.73\%. Similarly, significant improvements are observed in the audiovisual comparative QA: LAVisH+TWM improves by 10.08\%, DG-SCT+TWM by 13.56\% and LAST-Att+TWM by 10.92\%. TWM also leads to significant increases in overall average accuracy across all audiovisual tasks, with LAVisH+TWM improving by 6.35\%, DG-SCT+TWM by 8.58\% and LAST-Att+TWM by 5.01\%. By focusing on query-relevant segments and filtering out irrelevant content, TWM ensures that the model attends to the most informative parts of each modality, thereby improving cross-modal reasoning accuracy. This selective attention mechanism allows the model to better isolate critical elements within audiovisual streams, enabling it to reason about context-dependent relationships between different inputs. The fine-tuned multimodal focus highlights TWM's effectiveness in tasks requiring fine-grained comparisons, as it actively suppresses noise and enhances relevant signals.

\begin{table*}[htbp]
\centering
\resizebox{2.0\columnwidth}{!}{%
\begin{tabular}{@{}l|ccc|ccc|cccccc|c@{}}
\toprule
\multirow{2}{*}{{Method}} & \multicolumn{3}{c|}{{Audio-related QA}} & \multicolumn{3}{c|}{{Visual-related QA}} & \multicolumn{6}{c|}{{Audio\&Visual-related QA}} & \multirow{2}{*}{{Avg}} \\ 
 & {Count} & {Comp} & {Avg} & {Count} & {Local} & {Avg} & {Exist} & {Count} & {Local} & {Comp} & {Temp} & {Avg} & \\ \midrule
LAVisH \cite{Lin_2023_CVPR}& \textbf{83.82} & 58.19 & 75.72 & 82.81 & 81.73 & 82.30 & 73.26 & \textbf{73.45} & 65.64 & 64.26 & \textbf{60.82} & 67.75 & 73.28 \\
\textbf{LAVisH + TWM} & 79.22 & \textbf{70.59} & \textbf{76.91} & \textbf{84.52} & \textbf{84.10} & \textbf{84.31} & \textbf{78.66} & 65.21 & \textbf{73.05} & \textbf{74.34} & 57.24 & \textbf{74.10} & \textbf{74.42} \\
\textbf{Gain ($\Delta$)} & -4.60 & +12.40 & +1.19 & +1.71 & +2.37 & +2.01 & +5.40 & -8.24 & +7.41 & +10.08 & -3.58 & +6.35 & +1.14 \\ 
\midrule
DG-SCT  \cite{NEURIPS2023_af01716e} & \textbf{83.13} & 62.54 & 76.62 & 81.61 & 82.76 & 82.19 & 83.43 & 72.70 & 64.65 & 64.78 & \textbf{67.34} & 70.38 & 74.53 \\
\textbf{DG-SCT + TWM} & 80.05 & \textbf{72.66} & \textbf{77.20} & \textbf{87.77} & \textbf{85.24} & \textbf{86.35} & \textbf{88.69} & \textbf{87.21} & \textbf{76.06} & \textbf{78.34} & 59.82 & \textbf{78.96} & \textbf{79.22} \\
\textbf{Gain ($\Delta$)} & -3.08 & +10.12 & +0.58 & +6.16 & +2.48 & +4.16 & +5.26 & +14.51 & +11.41 & +13.56 & -7.52 & +8.58 & +4.69 \\ 
\midrule
LAST-Att \cite{liu2024tackling} & \textbf{86.03} & 62.52 & \textbf{79.44} & 84.12 & 84.01 & 84.05 & 76.21 & 75.23 & 68.91 & 65.60 & \textbf{60.60} & 69.04 & 75.44 \\
\textbf{LAST-Att + TWM} & 79.52 & \textbf{74.25} & 76.43 & \textbf{88.52} & \textbf{85.98} & \textbf{87.01} & \textbf{80.12} & \textbf{82.40} & \textbf{76.06} & \textbf{76.52} & 55.82 & \textbf{74.05} & \textbf{77.96} \\
\textbf{Gain ($\Delta$)} & -6.51 & +11.73 & -3.01 & +4.40 & +1.97 & +2.96 & +3.91 & +7.17 & +7.15 & +10.92 & -4.78 & +5.01 & +2.52 \\ 
\bottomrule
\end{tabular}
}
\vspace{-0.1cm}
\caption{Results of different models on the test set of MUSIC-AVQA 2.0 \cite{liu2024tackling}. \textbf{Bold} results indicate the better performance.}
\label{tab:tab1}
\end{table*}

\subsubsection{Video Captioning}

\begin{table}[!htb]
    \centering
    \resizebox{0.48\textwidth}{!}{%
    \begin{tabular}{@{}lccc@{}}
        \toprule
        \textbf{Method} & \textbf{ROUGE-L} $\uparrow$ & \textbf{CIDEr} $\uparrow$ & \textbf{SPICE} $\uparrow$ \\
        \midrule
        Git  \cite{wang2022git} & $24.51$ & $32.43$ & $13.70$ \\ 
        \textbf{Git + TWM} & $\mathbf{26.10}$ & $\mathbf{39.25}$ & $\mathbf{14.31}$ \\ 
        \textbf{Gain ($\Delta$)} & $+1.59$ & $+6.82$ & $+0.61$ \\
        \midrule
        AKGNN  \cite{hendria2023action} & $\mathbf{21.42}$ & $25.90$ & $\mathbf{11.99}$\\ 
        \textbf{AKGNN + TWM} & $21.33$ & $\mathbf{27.46}$ & $11.02$\\
        \textbf{Gain ($\Delta$)} & $-0.09$ & $+1.56$ & $-0.97$ \\
        \midrule
        CEN  \cite{nadeem2024narrativebridge} & $27.90$ & $49.87$ & $15.76$ \\ 
        \textbf{CEN + TWM} & $\mathbf{28.10}$ & $\mathbf{52.01}$ & $\mathbf{15.90}$ \\ 
        \textbf{Gain ($\Delta$)} & $+0.20$ & $+2.14$ & $+0.14$ \\
        \bottomrule
    \end{tabular}%
    }
    \caption{Test results of different models on MSR-VTT.}
    \label{tab:tab2}
\end{table}

\begin{figure*}[!htb]
\begin{center}
\includegraphics[width=1\linewidth]{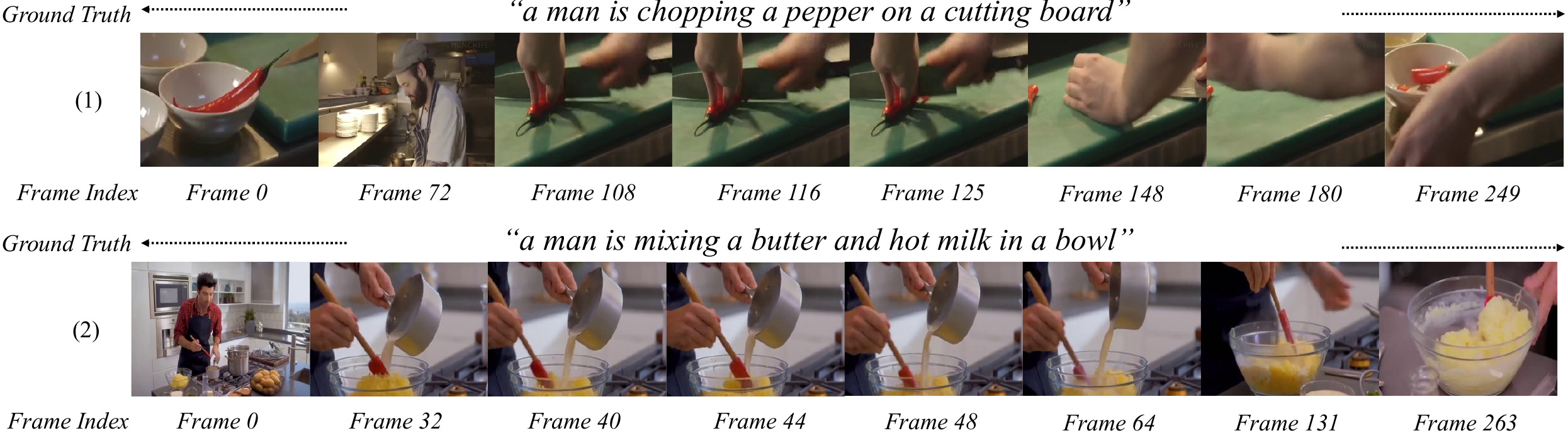}
\end{center}
\caption{TWM-searched frames for the video captioning task, integrated with AKGNN \cite{hendria2023action}. Two examples of frames searched by TWM are presented alongside their ground truth captions. The total number of frames per clip in MSR-VTT \cite{xu2016msr} ranges from 210 to 630. TWM-searched 8 frames effectively encapsulate the key visual information required to generate the ground truth captions.}
\vspace{0.2cm}
\label{fig:case_msr-vtt}
\end{figure*}

\paragraph{TWM enhances temporal coherence through selective attention to key events} 
In video captioning tasks (Table~\ref{tab:tab2}), TWM's selective attention mechanism significantly improves temporal coherence by focusing on key events and discarding irrelevant content. For example, Git+TWM achieves a 6.82\% improvement in CIDEr and a 1.59\% increase in ROUGE-L, highlighting the model's enhanced ability to generate coherent narratives that follow the flow of events. By retaining only the most contextually relevant audio-visual segments, TWM helps to avoid disjointed or fragmented scene descriptions, which is critical for accurately representing long or complex narratives.

\paragraph{TWM captures fine-grained scene transitions, enhancing descriptive richness} 
TWM also excels at capturing fine-grained details within scenes, allowing the model to generate richer and more informative descriptions. For example, CEN+TWM shows a 0.14\% improvement in SPICE, reflecting the model's enhanced ability to capture varied and accurate content in captions. In addition, Git+TWM shows a 0.61\% increase in SPICE, indicating an improved ability to capture changes in the audiovisual content of the video, such as changes in action or context. By dynamically updating memory with the most relevant visual and auditory elements, TWM ensures that critical details are highlighted, resulting in more contextually accurate and detailed output. This enhanced descriptive richness is essential in scenarios involving complex or rapidly changing scenes, where capturing semantic shifts in the narrative is key to generating informative captions.

\subsubsection{Video-Text Retrieval}

\begin{table}[ht]
    \centering
    \resizebox{0.48\textwidth}{!}{%
    \begin{tabular}{@{}lccc@{}}
        \toprule
        \textbf{Method} & \textbf{Recall@1 $\uparrow$} & \textbf{Recall@5 $\uparrow$} & \textbf{Recall@10 $\uparrow$} \\
        \midrule
        VINDLU  \cite{cheng2023vindlu} & $18.4$ & $36.4$ & $41.8$\\ 
        \textbf{VINDLU + TWM} & $\mathbf{20.5}$ & $\mathbf{38.2}$ & $\mathbf{44.6}$\\
        \textbf{Gain ($\Delta$)} & $+2.1$ & $+1.8$ & $+2.8$ \\
        \midrule
        TESTA  \cite{ren2023testa} & $21.5$ & $42.4$ & $50.7$ \\ 
        \textbf{TESTA + TWM} & $\mathbf{22.1}$ & $\mathbf{45.3}$ & $\mathbf{52.1}$ \\ 
        \textbf{Gain ($\Delta$)} & $+0.6$ & $+2.9$ & $+1.4$ \\
        \midrule
        MovieSeq  \cite{lin2024learning} & $25.8$ & $45.3$ & $50.3$ \\ 
        \textbf{MovieSeq + TWM} & $\mathbf{27.5}$ & $\mathbf{47.0}$ & $\mathbf{51.1}$ \\ 
        \textbf{Gain ($\Delta$)} & $+1.7$ & $+1.7$ & $+0.8$ \\
        \bottomrule
    \end{tabular}%
    }
    \caption{Test results of different models on CMD.}
    \label{tab:tab3}
\end{table}

\paragraph{TWM maintains retrieval performance across broader scopes through adaptive segment retention}
TWM's ability to enhance cross-modal alignment extends beyond immediate retrieval precision to broader retrieval tasks (Table~\ref{tab:tab3}). For instance, VINDLU + TWM achieves improvements of 2.1\% in Recall@1, 1.8\% in Recall@5, and 2.8\% in Recall@10. TESTA + TWM demonstrates gains of 2.9\% in Recall@5 and 1.4\% in Recall@10, showcasing TWM's capacity to retain relevant segments even in complex or diverse video datasets. Similarly, MovieSeq + TWM shows consistent improvements with a 1.7\% increase in Recall@1 and Recall@5, and a 0.8\% gain in Recall@10. These results indicate that TWM's memory update mechanism is flexible enough to adapt to a wide range of retrieval tasks. By selectively focusing on the most important audio-visual elements, TWM improves the model's ability to retrieve relevant content across larger candidate sets. This adaptive retention mechanism allows the model to effectively balance precision and scope, ensuring that both specific and broad retrieval queries benefit from TWM's selective attention and memory update strategies.

\subsection{The Impacts of Temporal Sequence Selection}
\label{twm_case_study}
To illustrate the effectiveness of TWM, we present a case study where TWM's search engine selects highly relevant frames based on the input of the language query (Figure \ref{fig:case_msr-vtt}).  Below, we provide a detailed analysis of how TWM ensures completeness of action representation, eliminates irrelevant noise, and optimizes model performance through selective frame reduction.

\paragraph{Completeness of action representation} 
TWM captures all the core stages of primary actions like chopping or mixing. For the chopping example, it captures the pepper being placed on the cutting board (Frames 0 and 72), the initiation of slicing (Frames 108 and 116), intermediate chopping stages (Frames 125 and 148), and the completion of the task (Frames 180 and 249). In the mixing example, it includes frames that depict the pouring of ingredients (Frames 40 and 44), the stirring process (Frames 48, 64, and 131), and the final state of the mixture. By covering all key moments, TWM provides the captioning model with a comprehensive understanding of the actions.

\paragraph{Elimination of irrelevant noise}
In cooking videos, various elements can distract from the main actions, such as the kitchen background, other utensils, or idle moments unrelated to the primary tasks like chopping. By selecting only the frames that focus on essential actions such as chopping or mixing in the examples, the TWM efficiently filters out these distractions, providing cleaner visual information for the captioning model. By minimizing distracting frames in the captioning inputs, the resulting descriptions of the events in the video become more accurate and useful.

\paragraph{Selective frame reduction for efficient model capacity utilization}
By selecting a limited number of informative frames that encompass all core stages, TWM optimizes memory usage and computational resources. It directs computational resources to the most relevant parts of the sequence by eliminating redundant or unrelated frames, thereby enhancing performance without overburdening the model. This selective retention not only allows for the precise capture of essential actions, but also facilitates efficient processing, significantly saving memory and speeding up the captioning model. Additional case studies on video-text retrieval and audio-visual question answering tasks are available in the Appendix~\ref{more_cases}.

\section{Conclusion}
We introduce Temporal Working Memory (TWM), a cognitive module designed to enhance multimodal foundation models' capabilities on complex video-audio-language tasks. Multimodal foundation models often struggle with dynamic multimodal input due to inefficient utilization of limited internal capacity. TWM addresses these limitations by \textit{(i)} maintaining memory states to process temporal multimodal sequences through selective segment retention; \textit{(ii)} modeling multi-scale temporal dependencies between video and audio inputs; and \textit{(iii)} extracting query-relevant information from rich multimodal data for efficient memory utilization. Our approach effectively improves the performance of nine state-of-the-art multimodal models on three temporal reasoning tasks: audio-visual question answering, video captioning, and video-text retrieval.

\section*{Limitations}
The effectiveness of TWM has been demonstrated on specific multimodal tasks and benchmarks, such as video captioning, question answering, and video-text retrieval. However, its generalizability to other in-the-wild domains remains unexplored. Extending TWM’s application to other more practical multimodal tasks would be future direction for real-world applications.

\section*{Ethical Considerations}
We examined the study describing the publicly available datasets used in this research and identified no ethical issues regarding the datasets. 

\section*{Acknowledgment}
This study is supported by the Department of Defense grant HT9425-23-1-0267.

\bibliography{ref.bib}

\appendix

\section{Baselines}
\label{baselines}
We evaluate nine MFMs reproduced with open-source code and pretrained weights:
\begin{itemize}[leftmargin=*]
\item \textbf{LAVisH} \cite{Lin_2023_CVPR}: LAVisH adapter uses a small set of latent tokens, forming an attention bottleneck that reduces the quadratic cost of standard cross-attention. The trained LAVisH module alongside Swin-v2 serves as TWM visual encoder.

\item \textbf{DG-SCT} \cite{NEURIPS2023_af01716e}: Adds cross-modal interaction layers to pretrained audio-visual encoders for adaptive extraction across spatial, channel, and temporal dimensions. All DG-SCT and Swin-T blocks serve as TWM visual encoder.

\item \textbf{LAST-Att} \cite{liu2024tackling}: Explores the interrelationships between audio-visual-text modalities. Swin-v2 serves as TWM visual encoder.

\item \textbf{Git} \cite{wang2022git}: Simplifies the architecture to a single image encoder and a text decoder under a unified language modeling task. The pretrained image encoder serves as TWM visual encoder.

\item \textbf{AKGNN} \cite{hendria2023action}: Introduces a grid-based node representation, where nodes are represented by features extracted from a grid of video frames. The trained graph neural network from AKGNN serve as TWM visual encoder.

\item \textbf{CEN} \cite{nadeem2024narrativebridge}: Utilizes independent encoders to capture causal dynamics and generate time-sequenced captions. The pretrained CLIP-ViT from CEN serves as TWM visual encoder.

\item \textbf{VINDLU} \cite{cheng2023vindlu}: Develops a stepwise approach for efficient VidL pretraining. The trained video encoder from VINDLU serves as TWM visual encoder.

\item \textbf{TESTA} \cite{ren2023testa}: Compresses video semantics by adaptively aggregating similar frames and similar blocks within each frame. The full video encoder blocks from TESTA serve as TWM visual encoder.

\item \textbf{MovieSeq} \cite{lin2024learning}: Through instruction tuning, MovieSeq enables a language model to interact with videos using cross-modal instructions. CLIP vision encoder from MovieSeq serve as TWM visual encoder.
\end{itemize}

\begin{figure*}[htb]
\begin{center}
\includegraphics[width=1\linewidth]{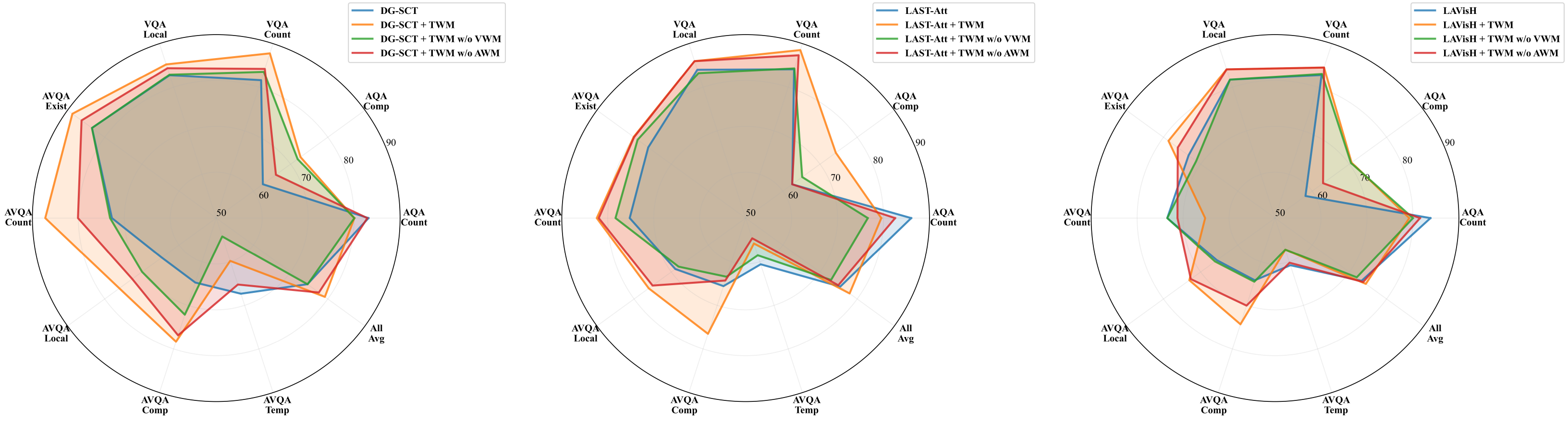}
\end{center}
\caption{Ablation studies on different components of TWM using MUSIC-AVQA v2.0 dataset~\cite{liu2024tackling}. We evaluate the effectiveness of two key components in TWM: Visual Working Memory (VWM) which refines temporal features in visual modality, and Audio Working Memory (AWM) which refines temporal features in audio modality. TWM w/o VWM indicates TWM is only applied to auditory modality while visual features retain the baseline model's distribution, and TWM w/o AWM indicates TWM is only applied to visual modality while maintaining the original auditory distribution.}
\label{fig:ablation_avqa}
\end{figure*}

\section{Ablation Studies}
To thoroughly investigate the effectiveness of two key components in TWM and analyze their relative contributions to model performance, we conduct comprehensive ablation experiments on the MUSIC-AVQA v2.0 dataset~\cite{liu2024tackling}. Specifically, we examine two critical variants: (1) \textbf{TWM w/o VWM}, where TWM is exclusively applied to the auditory modality while retaining the baseline model's original distribution for visual features, and (2) \textbf{TWM w/o AWM}, where TWM is solely applied to visual modality while maintaining the original auditory distribution.

As shown in Figure \ref{fig:ablation_avqa}, the ablation studies reveal several notable findings. First, Visual Working Memory (VWM) demonstrates substantial impact on both visual perception and cross-modal understanding. Taking DG-SCT+TWM as an example, the removal of VWM leads to significant performance degradation in VQA (-3.15\%) and AVQA (-8.39\%) average scores, indicating VWM's crucial role in capturing and refining temporal dependencies in visual features. Second, Auditory Working Memory (AWM) exhibits particularly strong effects on audio-related tasks, evidenced by a notable 4.67\% decrease in AQA performance when removing AWM from LAVisH+TWM, while its impact on visual tasks remains relatively minimal (e.g., -0.35\% in LAST-Att+TWM's VQA performance). 
Most notably, the complete TWM framework (AWM+VWM) consistently achieves promising performance across all evaluation metrics, with particularly impressive gains in cross-modal scenarios. For example, DG-SCT+TWM shows substantial improvements over its ablated variants, achieving an 8.39\% boost in AVQA average performance. These results empirically validate our hypothesis that AWM and VWM serve complementary functions in enhancing multimodal feature representations, which is essential for robust multimodal temporal reasoning.

\begin{figure}[htbp]
\begin{center}
\includegraphics[width=1\linewidth]{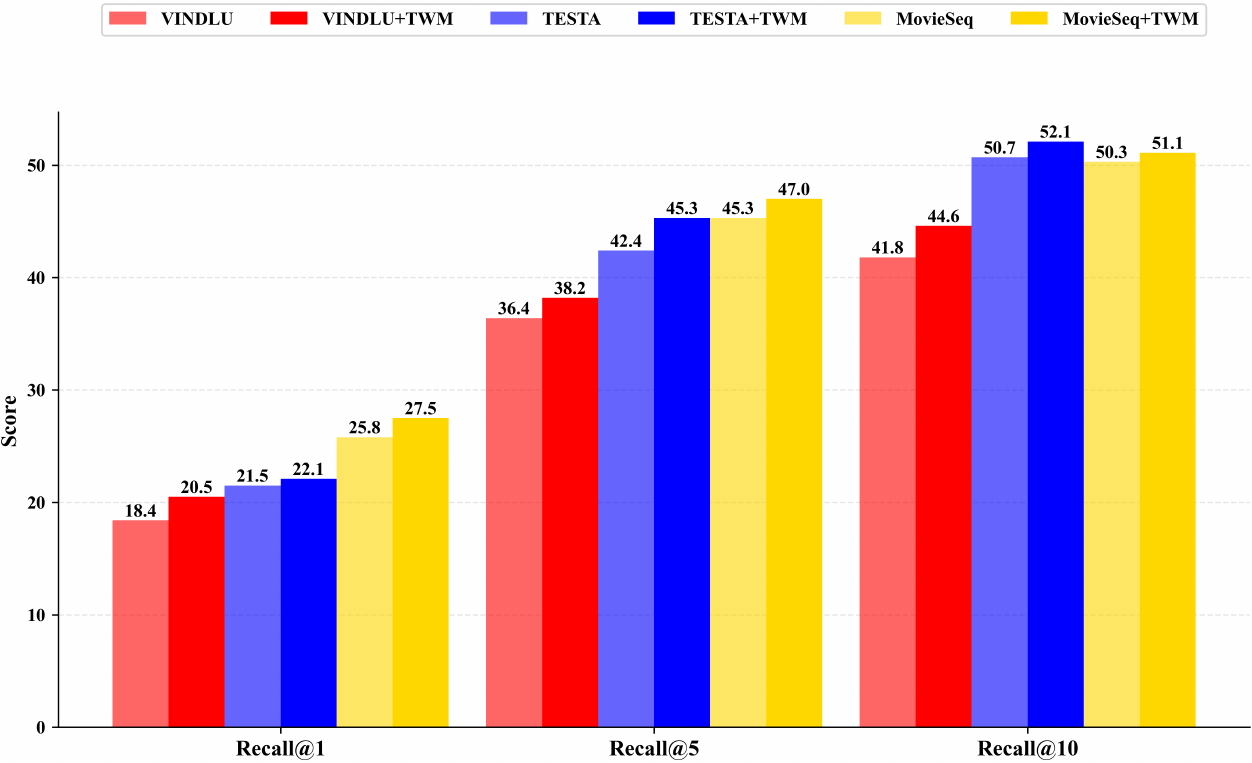}
\end{center}
\caption{TWM on video-text retrieval models using CMD dataset~\cite{bain2020condensed}. Lighter bars indicate baseline models while darker bars represent baseline models enhanced with TWM.}
\label{fig:ablation_retrieval}
\end{figure}

\begin{figure}[htbp]
\begin{center}
\includegraphics[width=1\linewidth]{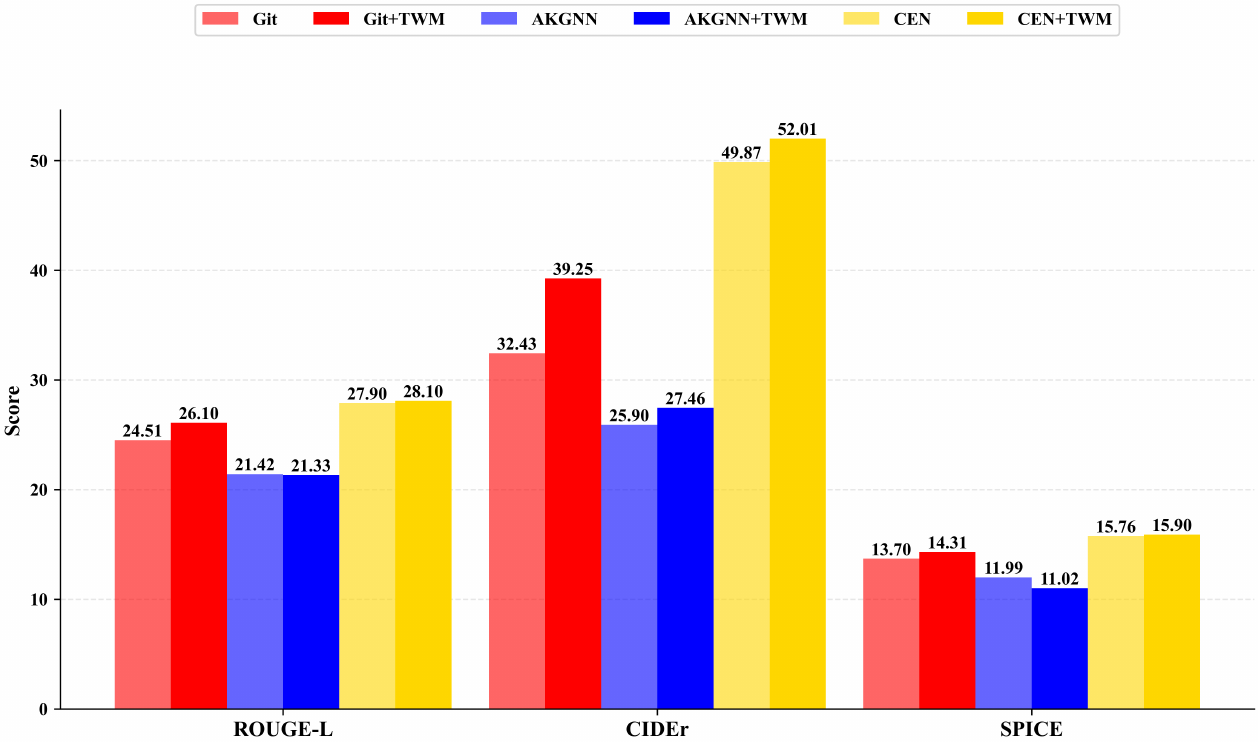}
\end{center}
\caption{TWM on video captioning models using MSR-VTT dataset~\cite{xu2016msr}. Lighter bars indicate baseline models while darker bars represent baseline models enhanced with TWM.}
\label{fig:ablation_caption}
\end{figure}

To further validate the generalizability of these findings across different tasks and datasets, we conduct additional quantitative evaluations illustrated in Figure \ref{fig:ablation_retrieval} and Figure \ref{fig:ablation_caption}. Figure \ref{fig:ablation_retrieval} presents a systematic performance evaluation of video-text retrieval models on the CMD dataset, where the comparative analysis demonstrates consistent performance improvements across multiple retrieval metrics after incorporating TWM. The empirical results from Figure \ref{fig:ablation_caption} further validate TWM's contribution through a comprehensive assessment of video captioning quality on the MSR-VTT dataset, where the evaluation spans multiple standard captioning metrics. These visualizations collectively substantiate TWM's capability to enhance model performance across different architectures and tasks, with particularly pronounced improvements observed in certain evaluation criteria.

\begin{figure*}[htb]
\begin{center}
\includegraphics[width=1.0\linewidth]{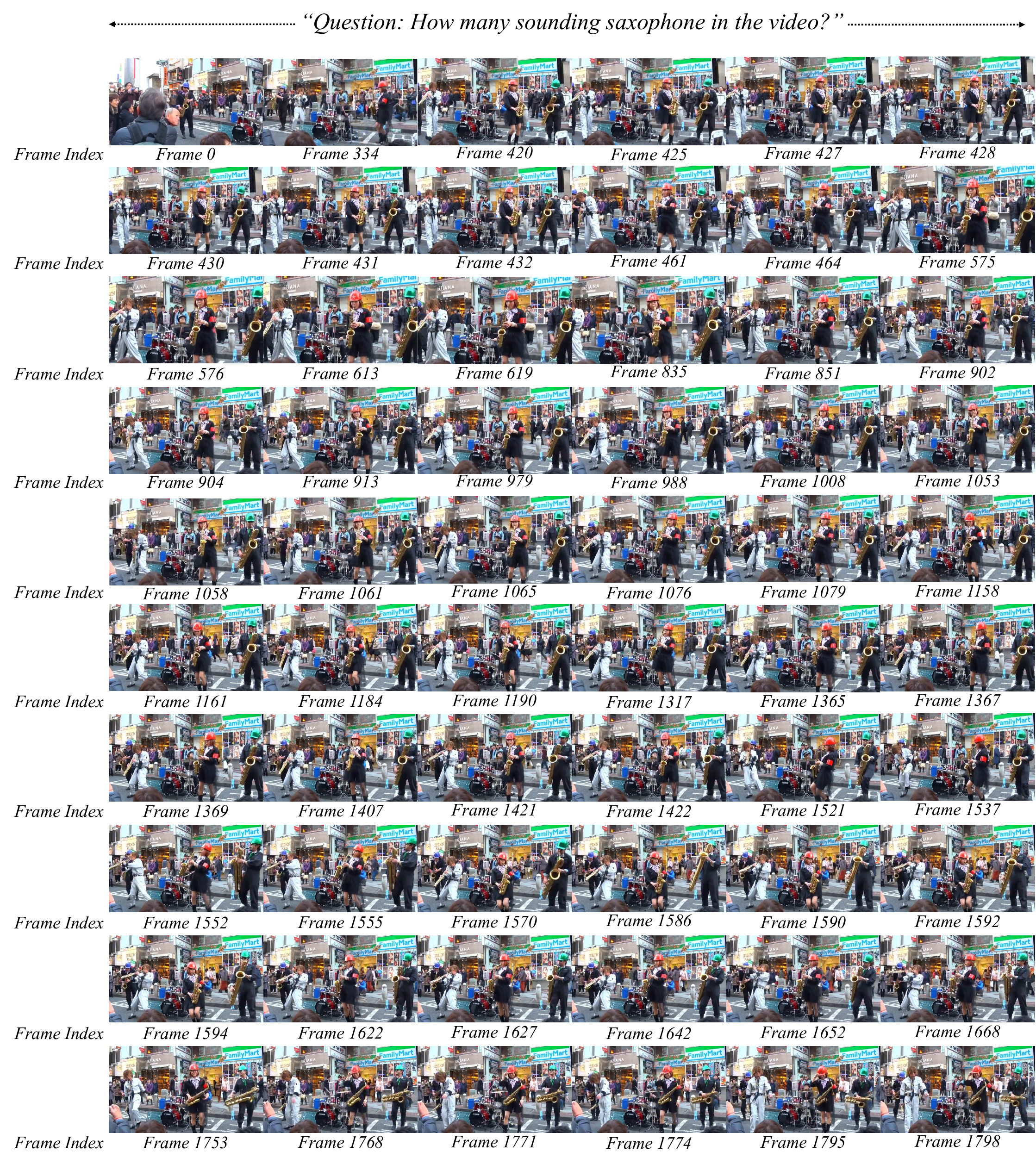}
\end{center}
\caption{TWM-searched frames with corresponding questions for audio-visual question answering, integrated with DG-SCT \cite{NEURIPS2023_af01716e} on the MUSIC-AVQA v2.0 dataset \cite{liu2024tackling}. The visualization demonstrates TWM's capability to search and select key frames that capture comprehensive audio-visual information necessary for answering the posed questions.}
\label{fig:case_dgsct}
\end{figure*}

\begin{figure*}[htb]
\begin{center}
\includegraphics[width=1\linewidth]{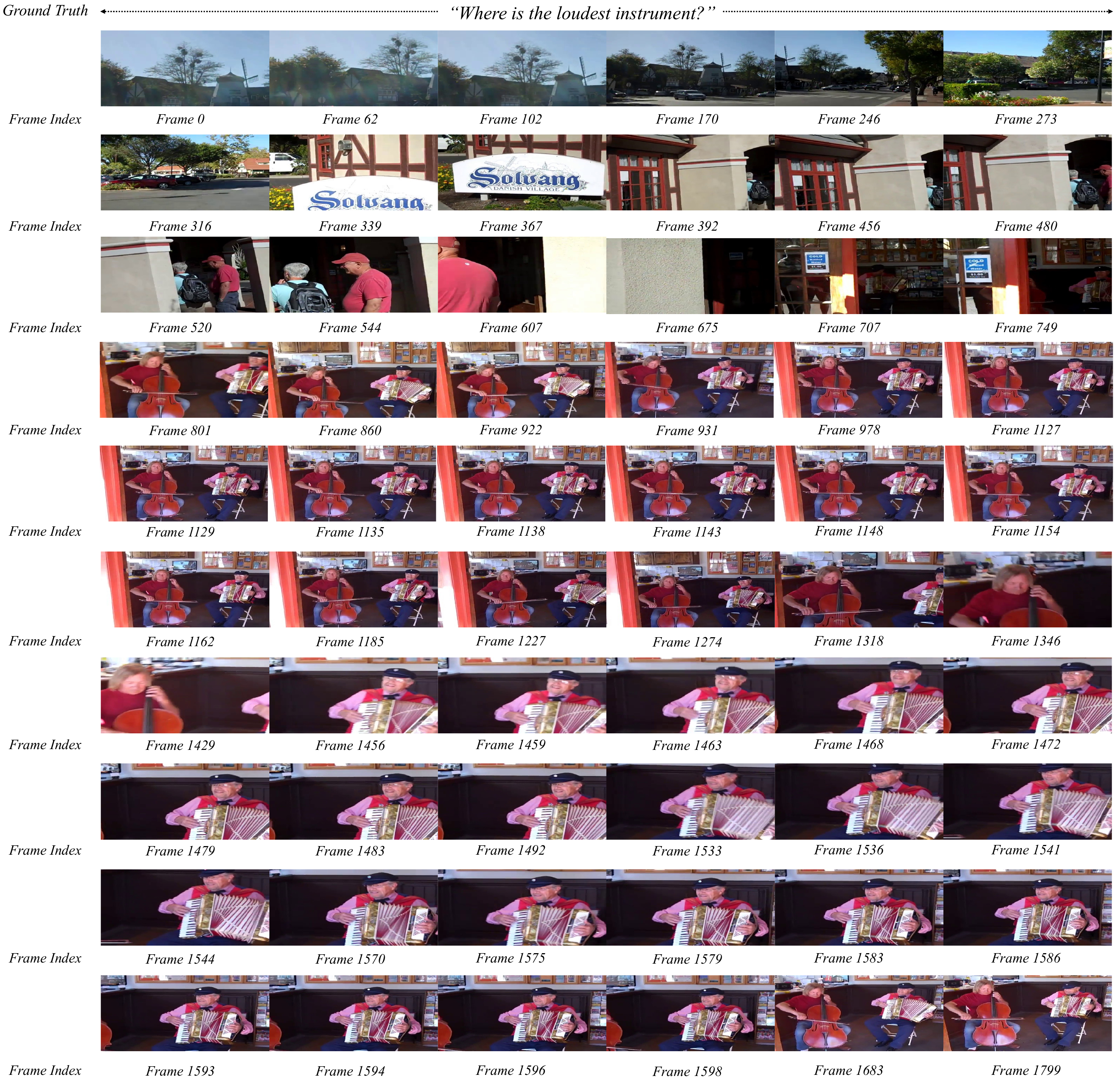}
\end{center}
\caption{TWM-searched frames with corresponding questions for audio-visual question answering, integrated with LAVisH \cite{Lin_2023_CVPR} on the MUSIC-AVQA v2.0 dataset \cite{liu2024tackling}. This visualization showcases TWM's capability in extracting frames that effectively capture the temporal-semantic information required to answer the given questions.}
\label{fig:case_lavish}
\end{figure*}

\begin{figure*}[htb]
\begin{center}
\includegraphics[width=1\linewidth]{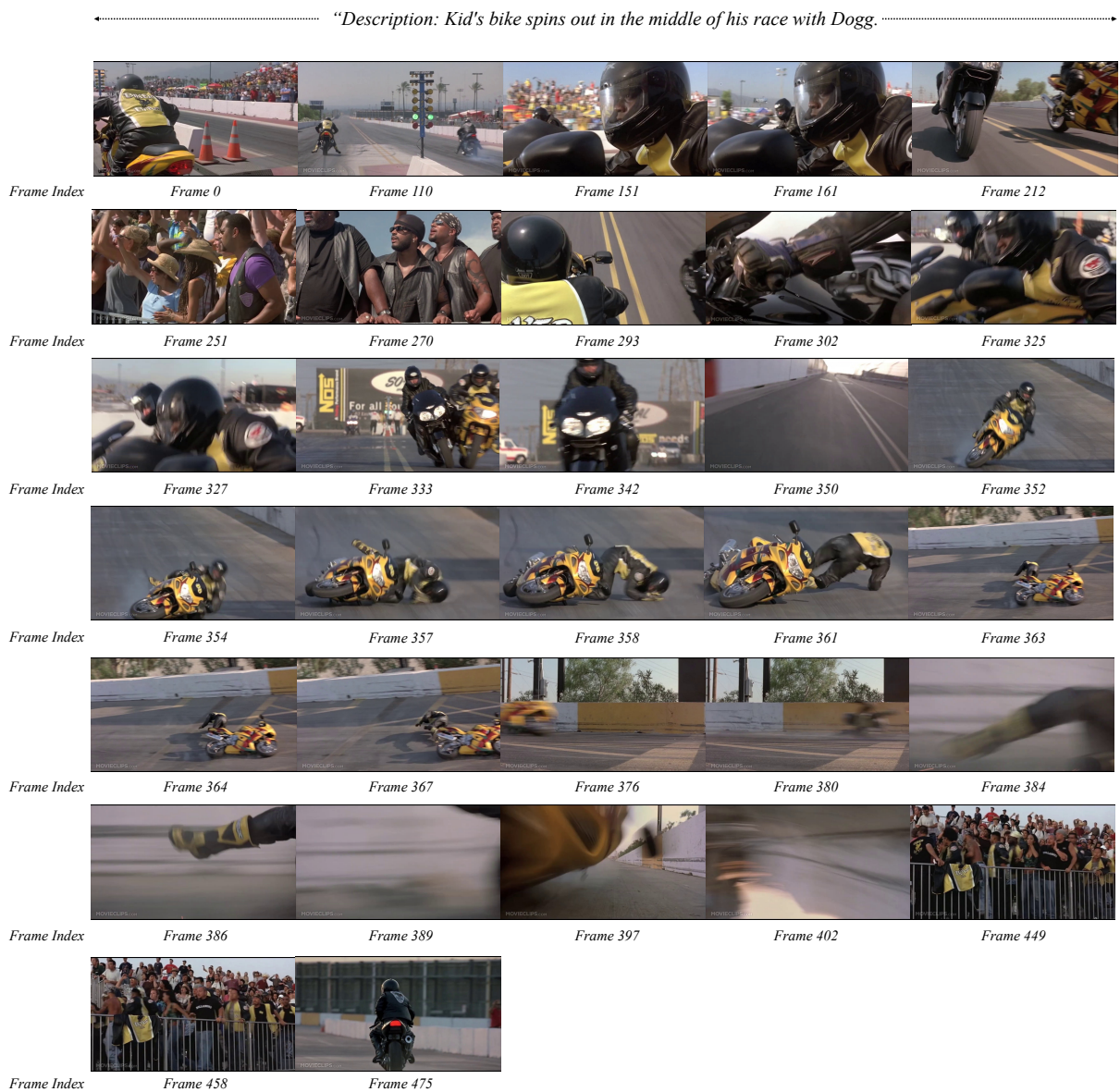}
\end{center}
\caption{TWM-searched frames with corresponding text description for video-text retrieval, integrated with TESTA \cite{ren2023testa} on the CMD dataset \cite{bain2020condensed}. The visualization demonstrates TWM's effectiveness in searching and selecting key frames that preserve the temporal-semantic information from the video sequence.}
\label{fig:case_testa}
\end{figure*}

\begin{figure*}[htb]
\begin{center}
\includegraphics[width=1\linewidth]{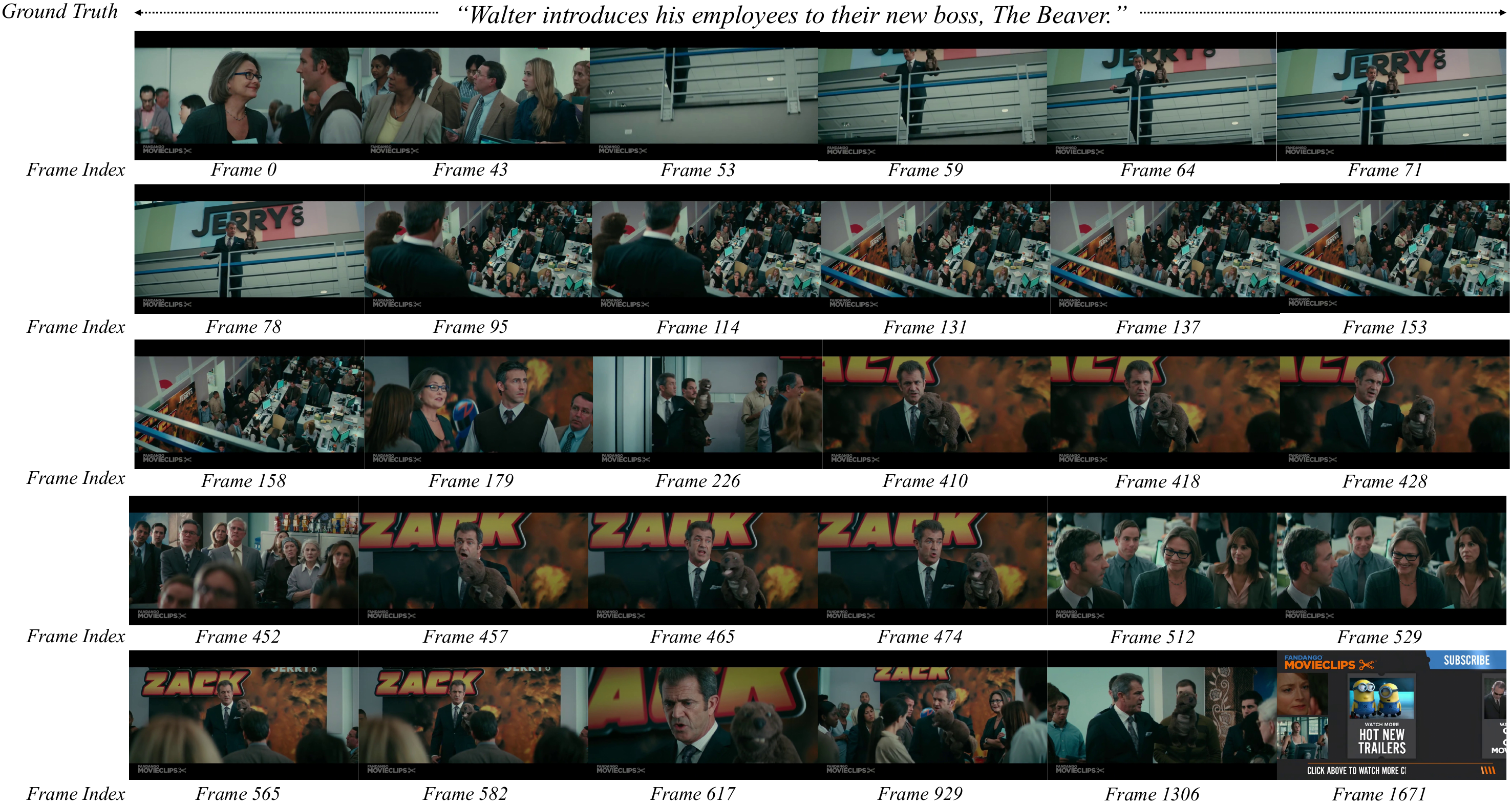}
\end{center}
\caption{TWM-searched frames with corresponding text description for video-text retrieval, integrated with MovieSeq \cite{lin2024learning} on the CMD dataset \cite{bain2020condensed}. This example illustrates TWM's capability to select visual frames that effectively capture the essential temporal dynamics and semantic information.}
\label{fig:case_movieseq}
\end{figure*}

\section{Additional Case Studies}
\label{more_cases}

In addition to Section \ref{twm_case_study}, we examine TWM's integration with different state-of-the-art models through four detailed case studies: crowd scene understanding, indoor performance analysis, dynamic action sequences, and social interaction comprehension. These cases highlight TWM's capability in frame selection, temporal coherence maintenance, and visual information preservation across diverse scenarios. 

\subsection{Crowd Scene Understanding}

Figure \ref{fig:case_dgsct} shows TWM's integration with DG-SCT \cite{NEURIPS2023_af01716e} on the MUSIC-AVQA v2.0 dataset \cite{liu2024tackling}, demonstrating its capabilities in handling complex crowd scenes in a street performance setting. The frame selection reveals TWM's systematic approach to managing dynamic crowd environments. In early frames, TWM establishes the overall scene context, capturing the spatial layout and crowd distribution. During the main sequence, it maintains tracking of individual performers while preserving their spatial relationships within the crowd. This strategic frame selection enhances DG-SCT's ability to understand complex crowd interactions while filtering out redundant or less informative frames.

\subsection{Indoor Scene Understanding}

Figure \ref{fig:case_lavish} demonstrates TWM's effectiveness when integrated with LAVisH \cite{Lin_2023_CVPR} on MUSIC-AVQA v2.0 for analyzing indoor performance scenes. Within this controlled environment, TWM exhibits precise frame selection that captures the key moments of performance. The selection strategy begins with establishing shots of the performance space, followed by carefully chosen frames that track the performer's movements. TWM's enhancement enables LAVisH to better handle indoor lighting conditions and maintain focus on the main subject despite varying camera angles and performer positions, demonstrating strong visual coherence in indoor settings.

\subsection{Dynamic Action Sequences}

Figure \ref{fig:case_testa} illustrates TWM's enhancement of TESTA \cite{ren2023testa} on the CMD dataset \cite{bain2020condensed} for handling fast-paced action in "Kid's bike spins out in the middle of his race with Dogg". The frame selection shows TWM's adaptive sampling strategy for dynamic events. The mechanism increases the sampling density during crucial moments of the action, particularly during the spin-out sequence. It maintains broader context through strategic selection of frames before and after the key event, enabling TESTA to better balance between capturing fine-grained action details and maintaining overall scene comprehension.

\subsection{Social Interaction Analysis}

The case in Figure \ref{fig:case_movieseq} showcases TWM's integration with MovieSeq \cite{lin2024learning} on the CMD dataset \cite{bain2020condensed} for capturing meaningful social interactions. Through the sequence "Walter introduces his employees to their new boss, The Beaver", TWM's frame selection effectively traces the progression of this social event. The initial frames establish the gathering context, showing the assembly of employees. The middle sequence frames capture key moments of the introduction, while later frames document audience reactions and interaction outcomes. This selection pattern enhances MovieSeq's ability to maintain narrative coherence through selective frame retention.

\subsection{Summary of Findings}

These real-world examples reveal several key aspects of TWM's operation in visual processing. The mechanism exhibits sophisticated understanding of scene dynamics, shown through its adaptive frame selection patterns across diverse environments. In crowd scenes with DG-SCT, it maintains multiple subject tracking while effectively filtering out redundant background information. For indoor venues with LAVisH, TWM demonstrates precise control over frame selection despite challenging lighting conditions and varying camera angles. In social interactions through MovieSeq, it shows awareness of narrative structure, selecting frames that preserve story progression, while in action sequences with TESTA, it employs dynamic sampling rates that respond to motion intensity and event significance.

TWM consistently achieves significant frame reduction while preserving essential visual information across different models and datasets. Its selection strategy ensures temporal coherence across selected frames, whether handling complex social scenes, dynamic actions, crowded environments, or indoor performances. This efficiency in frame selection, combined with maintenance of semantic continuity, highlights TWM's effectiveness as a visual memory management system.

\end{document}